\documentclass[transmag,letterpaper]{IEEEtran}
\newif\ifpeerreview

\peerreviewfalse

\newcommand{\paperID}{0021}

\usepackage[switch]{lineno}

\usepackage{cite}
\usepackage{url}
\usepackage{amsmath}
\usepackage{lipsum}
\usepackage{booktabs} % For formal tables
\usepackage{colortbl}
\usepackage{subfigure}
\usepackage{enumitem}
\usepackage[linesnumbered,algoruled,boxed,lined]{algorithm2e}
\usepackage{nicefrac}
\usepackage{xspace}
\usepackage{graphicx}

\definecolor{Yellow}{rgb}{1,1, 0.6}
\definecolor{Red}{rgb}{1, 0.6, 0.6}
\definecolor{Blue1}{rgb}{0, 0.6, 1}
\definecolor{Red1}{rgb}{0.5, 0.2, 0.2}
\definecolor{PaleYellow}{rgb}{0.8,0.8, 0}

\newcommand{\etal}{\textit{et al}.}
\newcommand{\ie}{i.e.\xspace}

\newcommand{\camone}{\emph{cam}\textsubscript{1}\xspace}
\newcommand{\camtwo}{\emph{cam}\textsubscript{2}\xspace}
\newcommand{\base}{\emph{base}\xspace}
\newcommand{\alt}{\emph{alt}\xspace}
\newcommand{\guide}{\emph{guide}\xspace}
\newcommand{\shortcite}[1]{\cite{#1}}

\ifCLASSINFOpdf
\else
\fi

\begin{document}

% activate line numbering to help with the review process
\ifpeerreview
  \linenumbers
  \linenumbersep 5pt\relax
\fi 

%
% paper title
% Titles are generally capitalized except for words such as a, an, and, as,
% at, but, by, for, in, nor, of, on, or, the, to and up, which are usually
% not capitalized unless they are the first or last word of the title.
% Linebreaks \\ can be used within to get better formatting as desired.
% Do not put math or special symbols in the title.
\title{Stereoscopic Dark Flash for Low-light Photography}
% \title{Stereoscopic Dark Flash Photography}
% \title{Dark Flash Stereopsis}
% \title{Dark Flash Stereoscopic Photography}
% \title{Dark Flash Stereo for Low-light Photography}

% author names and affiliations
% transmag papers use the long conference author name format.

\ifpeerreview
% Do not display author names for peer review since ICCP is double-blind. 
\author{Anonymous ICCP 2019 submission \\
Paper ID \paperID}
\else
\author{\IEEEauthorblockN{Jian Wang\IEEEauthorrefmark{1},
Tianfan Xue\IEEEauthorrefmark{2},
Jonathan T. Barron\IEEEauthorrefmark{2}, and
Jiawen Chen\IEEEauthorrefmark{2}}
\IEEEauthorblockA{\IEEEauthorrefmark{1}Carnegie Mellon University, Pittsburgh, PA 15232 USA}
\IEEEauthorblockA{\IEEEauthorrefmark{2}Google, Mountain View, CA 94043, USA}% <-this % stops an unwanted space
\thanks{ %Manuscript received December 1, 2012; revised August 26, 2015. 
Corresponding author: Jiawen Chen (email: jiawen@google.com).}}
\fi

% The paper headers
\ifpeerreview
\markboth{Anonymous ICCP 2019 submission ID \paperID}%
{}
\else
\fi

% As a general rule, do not put math, special symbols or citations
% in the abstract or keywords.
\IEEEtitleabstractindextext{%
\begin{abstract}
In this work, we present a camera configuration for acquiring ``stereoscopic dark flash'' images: a simultaneous stereo pair in which one camera is a conventional RGB sensor, but the other camera is sensitive to near-infrared and near-ultraviolet instead of R and B. When paired with a ``dark'' flash (\ie, one having near-infrared and near-ultraviolet light, but no visible light) this camera allows us to capture the two images in a flash/no-flash image pair at the same time, all while not disturbing any human subjects or onlookers with a dazzling visible flash. We present a hardware prototype of this camera that approximates an idealized camera, and we present an imaging procedure that let us acquire dark flash stereo pairs that closely resemble those we would get from that idealized camera. We then present a technique for fusing these stereo pairs, first by performing registration and warping, and then by using recent advances in hyperspectral image fusion and deep learning to produce a final image. Because our camera configuration and our data acquisition process allow us to capture true low-noise long exposure RGB images alongside our dark flash stereo pairs, our learned model can be trained end-to-end to produce a fused image that retains the color and tone of a real RGB image while having the low-noise properties of a flash image.
\end{abstract}

% Note that papers under review do not need to provide keywords
\ifpeerreview
\else
\fi
}

% make the title area
\maketitle

% To allow for easy dual compilation without having to reenter the
% abstract/keywords data, the \IEEEtitleabstractindextext text will
% not be used in maketitle, but will appear (i.e., to be "transported")
% here as \IEEEdisplaynontitleabstractindextext when the compsoc 
% or transmag modes are not selected <OR> if conference mode is selected 
% - because all conference papers position the abstract like regular
% papers do.
\IEEEdisplaynontitleabstractindextext
% \IEEEdisplaynontitleabstractindextext has no effect when using
% compsoc or transmag under a non-conference mode.

\begin{figure*}[h!]
	\center
	\includegraphics[width=0.98\textwidth]{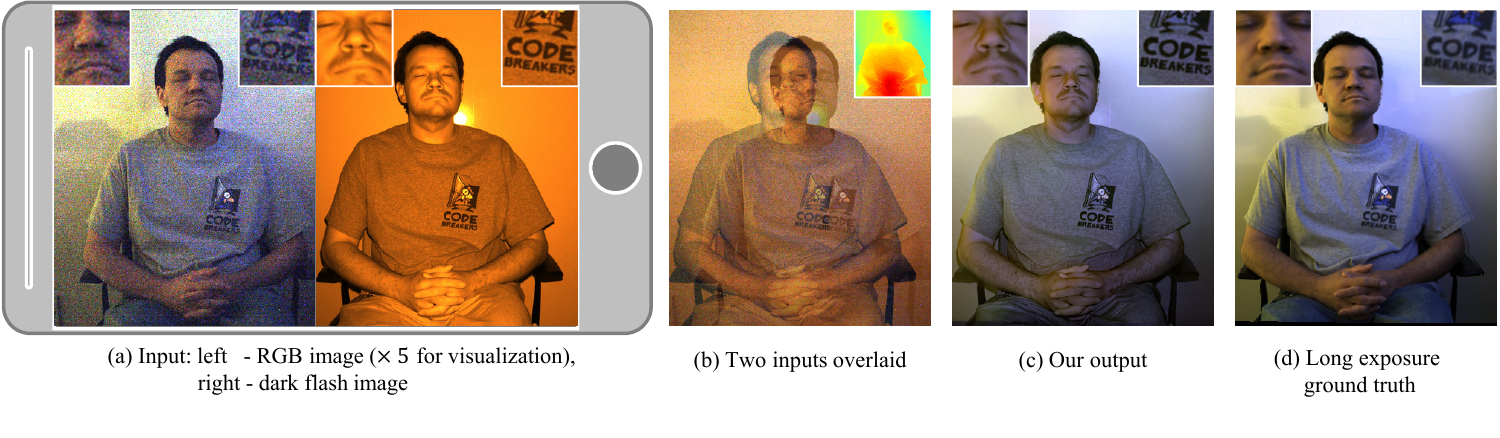}
    \vspace{-0.2in}
	\caption{Our stereo camera (a) simultaneously captures a conventional RGB image and an invisible ``dark flash'' image. We estimate stereo depth (b) and fuse them into a high-quality RGB result (c) that resembles a long-exposure ground truth RGB image (d).}
	\label{fig:teaser}
\end{figure*}

\section{Introduction}
% The very first letter is a 2 line initial drop letter followed
% by the rest of the first word in caps.
% 
% form to use if the first word consists of a single letter:
% \IEEEPARstart{A}{demo} file is ....
% 
% form to use if you need the single drop letter followed by
% normal text (unknown if ever used by the IEEE):
% \IEEEPARstart{A}{}demo file is ....
% 
% Some journals put the first two words in caps:
% \IEEEPARstart{T}{his demo} file is ....
% 
% Here we have the typical use of a "T" for an initial drop letter
% and "HIS" in caps to complete the first word.
%\IEEEPARstart{T}{his} demo file is intended to serve as a ``starter file''
%for submitting to the IEEE International Conference on Computational Photography. Example citation:~\cite{kopka-latex}. \lipsum[2-2]
% (should never be an issue)

\IEEEPARstart{T}{he} rise of mobile computing in the 21st century has caused photography to be a ubiquitous part of the human experience. But the small form factor of mobile phones necessarily limits the aperture size of the cameras that can be built into these devices, which in turn limits the amount of light that these cameras can detect. As a result, images taken by mobile devices in low light environments are often dominated by noise.

This issue can be ameliorated through conventional means, such as increasing the exposure time of the camera or using a flash, but these solutions have necessary drawbacks. Increasing exposure time allows more photons to be captured, but will induce a blur in the resulting photograph if the camera or subject moves --- barring the use of a tripod-mounted camera or a still life subject. Using a flash adds light to the scene but fundamentally changes the subject's appearance, often causing photos to look harsh or unnatural. In addition, using a flash may dazzle or otherwise disturb a human subject, or may transgress social norms in some circumstances.

Building on the idea of increasing exposure time, Hasinoff \etal~\shortcite{hasinoff2016burst} approximate a long-exposure image by capturing a burst of short-exposure images, and merging them together to obtain a lower-noise image. But this approach may still fail to reduce noise or eliminate blur in the presence of significant motion or very little light, and at best can only yield an SNR increase that is proportional to the square root of the number of images in the burst.
To address the sometimes-unattractive appearance of flash photographs, many researchers have explored capturing ``flash'' and ``no-flash'' image pairs and merging them to produce an image with the high SNR of the ``flash'' image, but with the attractive visual qualities of the ``no-flash'' image~\cite{petschnigg2004digital, eisemann2004flash}. Though it sometimes produces compelling results, because the flash/no-flash image pairs are taken at different times, this approach may fail in the presence of scene or camera motion. Additionally, a human subject would find it just as bothersome to be photographed by a flash/no-flash camera as they would a conventional flash camera.
To address the physiological and sociological problems associated with conventional flash photography, Krishnan and Fergus~\shortcite{krishnan2009dark} propose a ``Dark Flash'' that uses near-infrared (NIR) and near-ultraviolet (NUV) light which are invisible to the human eye. But their approach completely replaces the standard RGB sensor of a camera with another single sensor that conflates red with infrared, and blue with ultraviolet, and so they are entirely dependent on statistical correlations between visible and hyperspectral wavelengths to recover a visible-spectrum image. Though this correlation is strong, it is not deterministic, and so the inferred visible-spectrum image may contain significant artifacts. And because the sole sensor used by this camera configuration conflates visible with hyperspectral wavelengths, it cannot ``fall back'' to the RGB image as in the case of conventional flash/no-flash photography. Dark flash photography also inherits the vulnerabilities of flash/no-flash photography to camera or scene motion, as the two images of the pair are taken at different times.

% hence the name, dark flash, but needed to change the camera to be hyper-spectral responsive  and the no-flash image is no longer pure RGB image.

% Low-light imaging plays an important role in cell phone cameras, since people have huge demands to take photos at night time or entertainment places. 
% %
% Existing methods include applying high gain, using long exposure time, or using flash. However, applying high gain also boosts the noise level. Using long exposure time easily causes motion blur because of hand-shaking or scene movement.
% People also tried to capture a burst of photos with short exposure time to avoid motion blur and then merge them to obtain a high quality image \cite{hasinoff2016burst}, however, it is slow because of long exposure time and processing time. Using white flash can significantly boost the light level of the scene, but would dazzle or disturb people's eyes in an unpleasant way. Dilip and Fergus proposed to use near-infrared (NIR) and near-ultraviolet (NUV) flashes which are invisible to human eyes, hence the name, dark flash, but needed to change the camera to be hyper-spectral responsive \cite{krishnan2009dark} and the no-flash image is no longer pure RGB image.

\begin{figure*}[t]
	\center
    \includegraphics[width=1\textwidth]{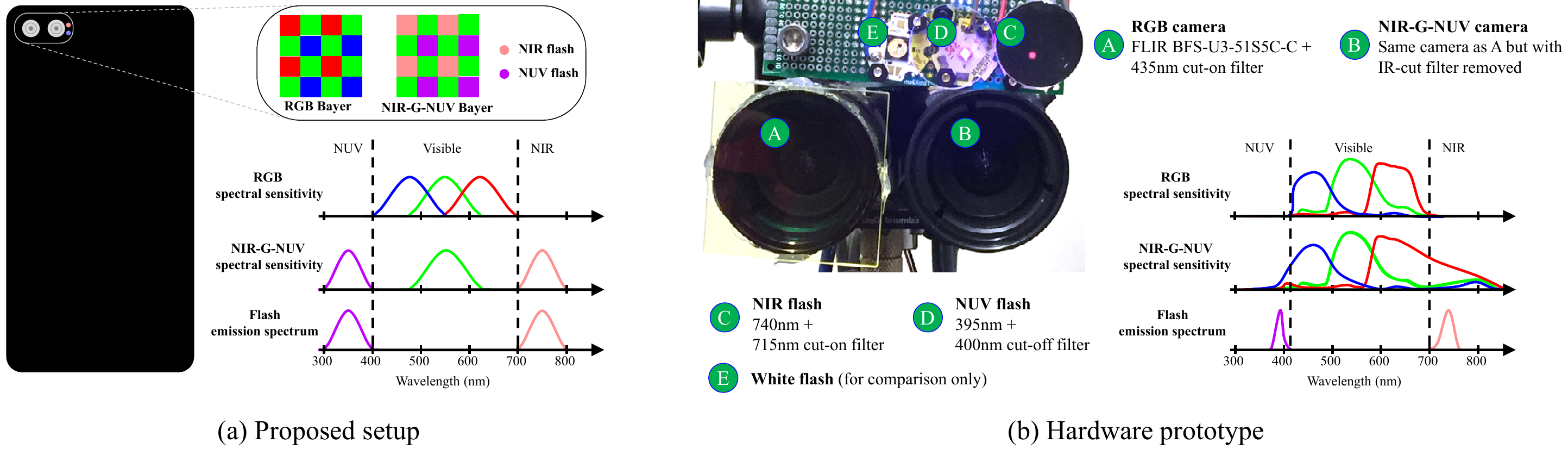}\\
	\caption{
    In (a) we show an idealized version of our proposed imaging setup: a mobile device with two cameras (one RGB, one NIR-G-NUV) and a dark flash, all with ideal spectral characteristics. In (b) we show our prototype camera system and its actual spectral curves.
    }
	\label{fig:setup}
\end{figure*}

Our approach, which we dub ``stereoscopic dark flash'', is an attempt to address some of the shortcomings of dark flash photography. Instead of using a single RGB camera with its IR/UV filter removed, we instead use two cameras: one standard RGB camera, and a second camera whose red and blue channels are replaced, making them sensitive to NIR and NUV respectively --- but insensitive to visible red and visible blue. Like in ``dark flash'' photography, our flash is limited to only NIR and NUV (see Figure~\ref{fig:setup} (a)). When taking a photograph, our camera rig fires the NIR-NUV flash and records an image from both the RGB and NIR-G-NUV cameras, all at the same time. Because the two cameras are at different physical locations, we must register the images to each other, which we do using their green channels (both of which are unaffected by the NIR-NUV flash and so appear similar). After a per-pixel alignment, we denoise the RGB image using the NIR and NUV channels of the flash image as a guide, which yields a low-noise image with natural RGB colors (see Figure~\ref{fig:teaser}).

As we will show, conventional methods for fusing a dark flash image with a noisy RGB image do not produce sufficiently high-quality results. The fused image may contain artifacts due to errors during registration, may have the characteristic harsh lighting of a conventional flash photograph, and may be corrupted by content not present in visible wavelengths. Thankfully, unlike in conventional dark flash photography, the presence of a standard RGB camera in our rig allows us to address this issue by collecting long exposure RGB images, which we use to train a neural network to regress from our naively fused RGB+NIR/NUV images to these ground truth RGB images. The resulting model learns to remove artifacts while retaining color information, and to modify the tonal content of the image to make it look more like a natural RGB image.

% To address this, we use our camera rig to collect a dataset of stereoscopic dark flash where for each image pair we additionally capture a long-exposure image 

% We propose dark flash stereo camera pairs for low-light imaging, in which in addition to the existing RGB camera, we add an NIR-G-NUV camera and NIR and NUV flashes. By one shot, the system acquires an RGB image and an NIR-G-NUV image. We align the two images by their G-channel, and then denoise the RGB image guided by the flash image. In the denoising step, we allow flash content to be introduced into denoised result so that denoised image has many details, but tone might be changed. We finally apply convolution neural network to adjust the tones to make denoised image look natural. 

The advantages of our proposed setup are many:
\begin{enumerate}
\item Our technique retains all of the value contributed by dark flash photography: low noise images can be recovered, without having to dazzle or disturb any human subjects or bystanders.
\item Unlike burst photography, flash/no-flash, and dark flash photography, our two images are acquired at the exact same time. This not only allows for a responsive and low-latency user experience, but also means that our system is robust to camera or scene motion.
\item Like in conventional flash/no-flash photography, our rig directly acquires a conventional RGB image. Contrast this with dark flash photography, in which the no-flash image is \emph{not} an RGB image, but instead contains both red and NIR wavelengths in the ``red'' channel and both blue and NUV wavelengths in the ``blue'' channel. This means that in well-lit environments in which noise is not an issue, we can produce a high-quality image by simply returning the observed RGB image. Moreover, in the case of some failure either during registration or image fusion, our setup can always degrade gracefully back to the observed RGB image if need be.
\item Unlike dark flash (but similarly to a flash/no-flash or burst photography setup) our setup allows for the collection of long-exposure ``ground truth'' RGB images, which can then be used for learning.
\item By constructing the spectral response curves of our cameras and flash such that the green channels are identical and have no overlap with our flash, conventional stereo techniques (which, naturally, expect input images to look similar) can be used to recover a reliable depth map. The depth maps we recover may also be useful for other photographic purposes such as background defocus \cite{barron2015fast}.
\end{enumerate}

However, our technique does have some limitations and costs that are worth considering:
\begin{enumerate}
\item The intensity of our flash, like any light, necessarily decreases with distance, and so our approach will not provide a benefit in distant scenes.
%low-light environments.
\item Because we rely on a stereo algorithm to align our input images, our output image may contain some artifacts around occlusions or other image regions where correspondence is difficult to compute.
\item Compared to a standard stereo rig, our setup has slightly less information with which to estimate disparity, as we use only the green channel shared by our two sensors while a standard stereo setup can use all three RGB channels. In practice, we observe that the drop in depth quality when using green instead of RGB appears to be small.
\item Unlike burst, flash/no-flash, or dark flash photography, our approach requires two cameras instead of one, which increases the cost of manufacturing and calibrating such a device.
\end{enumerate}

\section{Related Work}

Modern camera sensor design has been the focus of decades of research and engineering, much of which is well beyond the scope of this review. Conventional mobile cameras are constructed by placing a Bayer color filter array (CFA) \cite{bayer1976color} in front of a CMOS sensor that is sensitive to light in the range of $300$ to $1000$ nanometers. A Bayer filter is composed of repeated ``quads'' containing four pixels, each of which sits behind a color bandpass filter that eliminates all but red, green, or blue light. After the image is captured, a demosaicking algorithm~\cite{malvar2004high} interpolates an estimate of the two colors at each pixel that were not directly observed, resulting in a complete color image. Note that, by design, a Bayer filter approach discards approximately two thirds of all incoming light.
Though these CFAs are typically designed to filter out all but visible light, they have some leakage in the NIR range that necessitates an NIR cut-off filter, which is usually placed on the lens of the camera. The final spectral sensitivity curves for the sensor pixels of different camera brands are well summarized in~\cite{jiang2013space}. While CMOS sensors have benefits over CCD sensors in terms of power efficiency and readout speed, they require additional space for per-pixel readout and amplifier circuits, and therefore have less area that can be used for observing incident light \cite{litwiller2001ccd}. These issues, compounded by the aforementioned limits on the size of mobile devices and the commonality of low-light environments, result in a camera whose images are often noisy and therefore benefit from a denoising algorithm or additional information that can be used to reduce noise.

% self denoising
Image denoising has been the subject of significant research, with many techniques using a single image as input, such as BM3D \cite{dabov2007image}, sparse coding \cite{elad2006image}, low-rank factorization \cite{gu2014weighted}, or modern deep learning based methods \cite{chen2018learSeeDark}. These methods are generally computationally expensive, and are necessarily limited in their ability to recover details in the presence of overwhelming noise. Performance can be improved by using a burst of images to denoise a single image \cite{hasinoff2016burst, fast-burst-images-denoising, Dabov2007, heide2014flexisp}, though these approaches require computing a correspondence across images or some technique for being invariant to this correspondence problem, which can be problematic in the presence of significant camera or scene motion.

Instead of acquiring a noisy image and then attempting to remove that noise, one can instead adjust the imaging conditions to capture a less noisy image. As mentioned previously, increasing the exposure time of the camera reduces noise, but results in blurring artifacts in the presence of scene motion or camera shake. This motion blur can be removed through algorithmic means \cite{Fergus06}, but this ``deblurring'' problem is itself difficult and underconstrained, and arguably no easier than the denoising problem that is being circumvented.
Alternatively, one could reduce noise by increasing the amount of illumination in the scene, through the use of a flash. Flash photographs tend to have an unpleasantly harsh and unnatural appearance, but this can be reduced by merging a flash photograph with a no-flash image~\cite{petschnigg2004digital, eisemann2004flash}. But even if the flash/no-flash problem is solved, many people still find the bright and dazzling white flash of a camera to be annoying or otherwise disruptive. Dark flash photography \cite{krishnan2009dark} avoids this problem by using NIR and NUV flashes, and modifying the camera to be NIR/NUV-sensitive by removing the IR/UV-cut filter, though this system has its own drawbacks, as explained previously.

% software side: how to do guided denoising (and tone mapping)
The flash/no-flash imaging strategy presents the question of how to best combine the high fidelity of the flash image with the more pleasing aesthetic qualities of the no-flash image, and this question has received a significant amount of attention.
Early approaches \cite{petschnigg2004digital,eisemann2004flash} use joint bilateral filtering to produce a ``detail'' layer from the flash image that is then propagated to the no-flash image. Other edge-aware filters, such as the guided filter~\cite{he2010guided} can be used similarly.
Dark flash photography~\shortcite{krishnan2009dark} merges its two images using an optimization framework that assumes the gradient of the denoised result should be similar to the gradient of the flash image. Though image gradients are often strongly correlated across different wavelengths, the occasional variations that do occur can cause algorithms that depend on this assumption to fail.
To address this issue, Shen et al.~\shortcite{shen2015multispectral} explicitly model the structural divergence across wavelengths as a \emph{scale map} --- the ratio between the gradient maps of the flash and no-flash images, which they jointly estimate alongside a denoised image.
% The denoised result may introduce gradients from the flash image which is good when the gradients are from scene's texture, and bad when from the flash shadows and specularities and tone is not natural.
Similarly, mutually guided image filtering \cite{guo2017mutually} also propagates information across disparate wavelengths through a joint estimation process.  
% In contrast, muGIF \cite{guo2017mutually} also respects the spectrum difference but does not introduce new content from flash image.
Though these techniques work well, their hand-engineered nature means that they often propagate gradient information that has undesirable tonal properties from the flash image that would not be present in a long-exposure image. For this reason, we build upon \cite{shen2015multispectral} but augment it with a neural network that has been trained  to remove the unwanted tonal and color properties of the dark flash.

% This technique successfully propagates gradient information across the two images, but may still fail in the absence of any guiding information in the dark flash image, and may propagate unwanted tonal properties from the flash image.

\section{Proposed Setup}

Our proposed imaging configuration consists of a conventional RGB camera (\camone), an NIR-G-NUV camera (\camtwo), an NIR flash, and an NUV flash. In Figure~\ref{fig:setup}(a), we illustrate the envisioned use of our system in a cell phone, alongside the Bayer pattern of the two cameras and the idealized spectral response curves for both sensors' micro-filters and the corresponding flashes.
This idealized camera captures a single shot by simultaneously firing both dark flashes while exposing both sensors.
The flash is invisible to both the human eye as well as \camone, but is visible to \camtwo thereby allowing it to record a low-noise flash image in low-light environments.
Because the two cameras have different positions, merging the two images requires solving for a per-pixel mapping. We do this by using the green channels of the two images (which, because the green curves of the two cameras are matched, look similar) to estimate a dense stereo depth map, and use this depth map to help in merging the two images together to produce a single high-quality RGB result.

\subsection{Our prototype}
We built our prototype camera system using off-the-shelf components that approximates our proposed configuration (Figure~\ref{fig:setup}(b)). In this section, we detail our design choices, the practical limitations caused by our selection of hardware, and how these choices affect our dataset capture strategy. The supplement contains details on hardware components and their specifications.

\begin{figure}[t]
	\center
    \includegraphics[width=\linewidth]{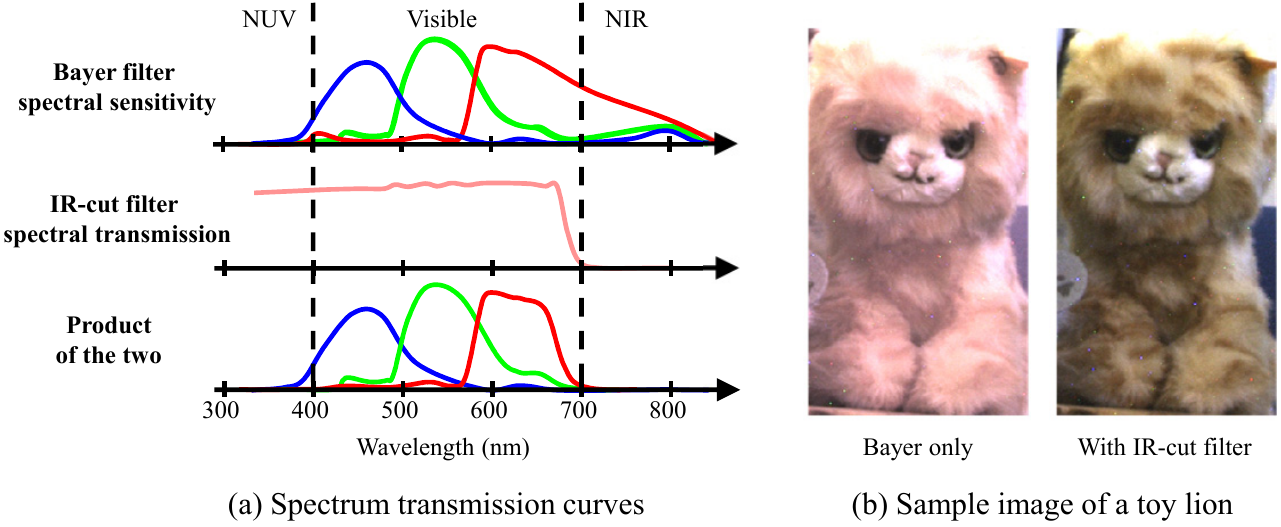}\\
	\caption{(a) Spectral transmission curves of Bayer filters, low-pass NIR filter, and their product. (b) Sample images with and without the NIR filter.}
	\label{fig:IRcut_effect}
\end{figure}

We chose to match the sensor sizes and lenses of our two cameras, giving us images with the same field of view and resolution. This was done to minimize the difficulty of performing stereo registration, and to prevent artifacts in the final merged image that may result from mismatched FOVs causing the observed areas of the two images to be significantly different.
We selected FLIR sensors that are sensitive to RGB, NIR, and NUV light, and we chose lenses that allow NIR through NUV wavelengths to pass through, and that also produce little chromatic aberration when focusing all relevant wavelengths.

Our left camera (\camone) is a standard RGB camera, to which we add a UV filter on its lens (which was already equipped with an NIR filter) to ensure that it is only sensitive to visible light.
Our right camera (\camtwo) is an NIR-G-NUV camera. Because it is difficult to construct a camera that is sensitive to different wavelengths than \camone but is otherwise physically identical, we instead build \camtwo by modifying an RGB camera.
Recall that Bayer micro-filters are not true band-pass filters, as they transmit significant amounts of light outside the visible spectrum (see Figure~\ref{fig:IRcut_effect}). This means that if we remove the NIR-cut filter from the lens, we can make an RGB camera sensitive to both NIR and NUV. Unfortunately, this also means that its green channel will receive a non-negligible amount of NIR that is sometimes problematic for our stereo algorithm.
To address this issue, our dataset consists exclusively of indoor scenes, which allows us to minimize the amount of ambient IR and UV light that is present during imaging.
Still, despite our best efforts, we noticed that images from \camtwo have a slight red tint and are slightly blurrier due to NIR contamination and chromatic aberration.

To demonstrate the feasibility of our algorithms despite whatever practical issues with our hardware we may have, during capture we simulate an ideal shot by acquiring bursts of images for each scene. Our bursts rapidly interleave shots with flash off and flash on. We compute stereo correspondence on the flash-off frames (where the green channels are uncontaminated by NIR), but use this estimated depth map to warp the flash-on frames. This burst capture also let us benchmark our method against burst denoising algorithms as an alternative strategy for producing low-noise RGB images.

Our difficulty in constructing a physical prototype that exactly matches our proposed setup may prompt the reader to question if it is indeed possible to manufacture pure sensor-level R, G, B, NIR, and NUV Bayer filters. Indeed, Spooren~\etal~\shortcite{spooren2016rgb} show that it is possible to construct a compact, low-cost RGB-NIR camera --- albeit one that is difficult to procure --- using pixel-level monolithic integration of traditional absorption-based RGB color filters with NIR-pass and NIR-cut filters implemented using Fabry-P\'erot interference. By using Fabry-P\'erot interference filters and mosaicking different filters to select different wavelengths, others have successfully implemented Bayer hyperspectral cameras \cite{tack2012compact,geelen2015tiny} which exceed our setup's requirements.

\section{Burst Dataset}

% describe how the burst is captured:
% AE algorithm, burst implementation
% different flash time is for motion blur

% figure: one figure to show a group of data, to show how the data look like in the dataset

\begin{figure*} 
	\center
	\includegraphics[width=1\textwidth]{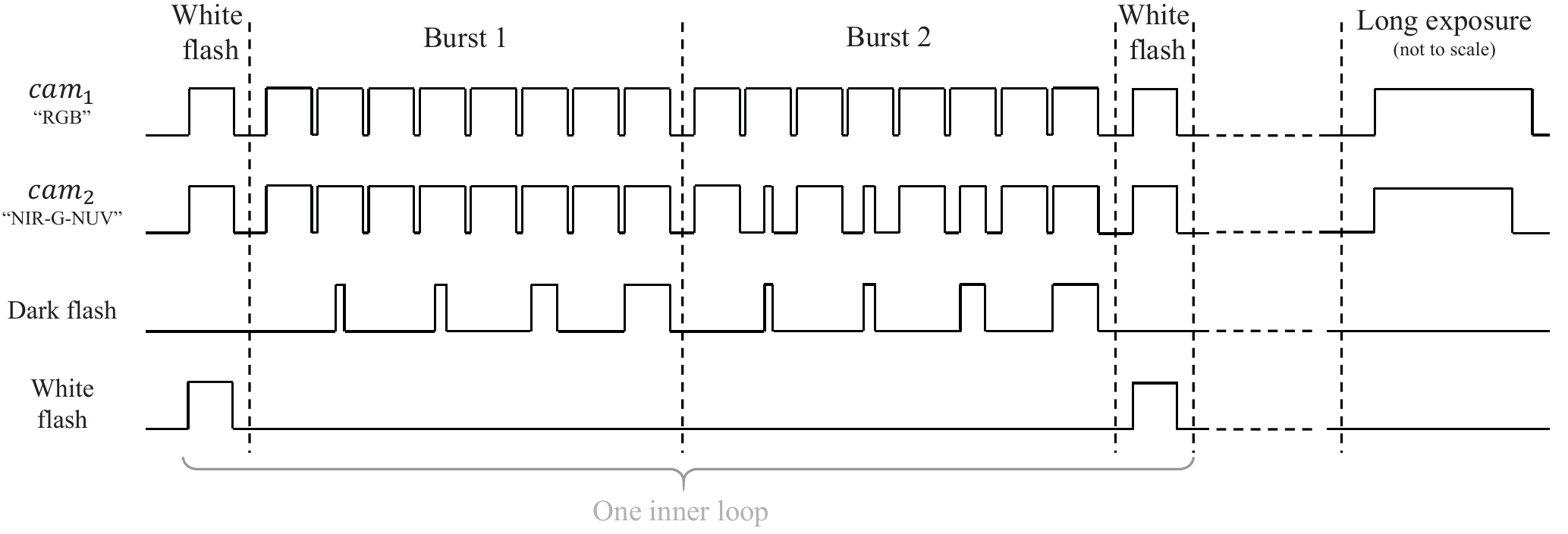}\\
    \vspace{-5pt}
	\caption{Our data collection strategy, which includes white flash, two bursts under varying length dark flashes, and two no-flash long exposure images.}
	\label{fig:dataset_strategy}
\end{figure*}

To facilitate our experiments we collect a dataset of 121 scenes, which will be used for training our model and to benchmark our model's performance against baseline techniques. This dataset collection procedure is designed to be more general and flexible than is needed for this work, in the hopes that a large and rich dataset of this sort may be a useful resource for future research.

Because we acquire bursts of images, and because the problem of image registration is difficult and somewhat out of the scope of this work, our strategy is to minimize differences across images due to anything other than flash (\ie, motion). To this end, all images are captured using a tripod. 30 of our scenes are of static environments, and the remaining 91 scenes contain human subjects. These subjects were told to hold still during acquisition, though our human-subject scenes do generally contain small amounts of motion.

For each scene, we first run a simple automatic exposure algorithm (see the supplement) to estimate an appropriate exposure time $T$ as well as gains for the two cameras. We then capture a collection of bursts, where for each burst we vary one property of our acquisition setup (see Figure~\ref{fig:dataset_strategy}). Our acquisition procedure is described in Algorithm~\ref{alg:burst}.
\begin{algorithm}[b]
\caption{Our Burst Collection Procedure}\label{alg:burst}
\For{$t \in [T, \nicefrac{T}{3}, \nicefrac{T}{5}, \nicefrac{T}{7}] $}{
  \For{ $\mathrm{flash} \in [\mathrm{NIR}, \mathrm{NIR} + \mathrm{NUV}]$}{
  	Capture 1st still image with white flash on\;
	Capture burst 1\;
	Capture burst 2\;
	Capture 2nd still with image white flash on\;
  }
}
Capture long exposure ground truth\;
\end{algorithm}

% \indent --- for $t = T, T / 3, T / 5, T / 7 $ \\
% \indent \indent --- for flash = \{NIR-only, NIR and NUV\} \\
% \indent \indent \indent \indent Capture 1st still with white flash on; \\
% \indent \indent \indent \indent Capture burst 1; \\
% \indent \indent \indent \indent Capture burst 2; \\
% \indent \indent \indent \indent Capture 2nd still with white flash on \\
% \indent --- Capture long exposure ground truth.
For \camone, we simply capture a uniform burst with exposure time $T$. For \camtwo, recall that it is an approximation of an ideal NIR-G-NUV sensor. We therefore collect two bursts with different \camtwo exposure times when the dark flash is on to assess the tradeoff between motion and NIR contamination in stereo registration. Burst 1 maintains a uniform exposure time, thereby ensuring equivalent motion blur in the two images, but \camtwo's green channel records ambient NIR in addition to that from the flash. In Burst 2, we match \camtwo's flash-on exposure times with that of the flash. This minimizes NIR contamination at the expense of mismatched motion blur. In practice, we found little difference between the two since our scenes are largely static. We use burst 1 for all of our results.

\begin{figure*} 
	\center
	\includegraphics[width=1\textwidth]{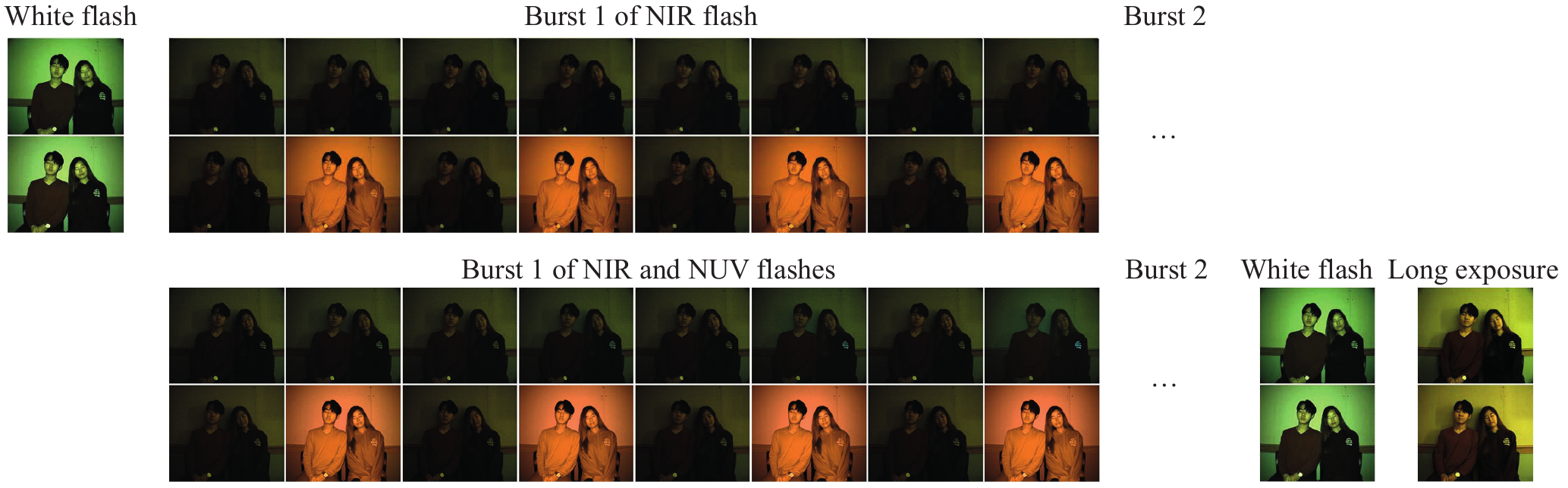}\\
    \vspace{-7pt}
	\caption{An example scene from our dataset. We show a series of image pairs where the upper image is from $cam_1$ (RGB) and the lower image is from $cam_2$ (NIR-G-NUV). Burst 2 is omitted for space, as it resembles burst 1. The flash images with different exposure time use different analog gains such that they have same level of brightness. These linear images have not been white balanced, hence their green or orange appearance. %Please see the supplement for more examples from our dataset.
    }
	\label{fig:dataset_sample}
\end{figure*}

For each exposure time and flash combination, we bookend the two bursts with a white flash image. Finally, we also capture a long-exposure ground truth image.
All images are captured in 16-bit Bayer raw. Figure \ref{fig:dataset_sample} shows one example of the many images acquired for one scene in our dataset. Note that these bursts are used only for training and evaluation, and that this burst acquisition procedure is \emph{not} necessary for acquiring test-time photographs using our camera rig.
%\jc{At test time, we require only two stereo pairs (flash off and flash on) as input, represented by two adjacent columns in Figure~\ref{fig:dataset_sample}. The ideal setup would require only a single pair.} \jc{This is some complicated foreshadowing. Can we just defer to seciton 5?}

Though our acquisition process is somewhat complicated, the resulting data we acquire has a number of useful properties:
\begin{itemize}
 \item The long-exposure RGB images can be used as ground truth for training and evaluating models.
  \item Because we acquire an RGB burst, we can directly compare our results against a standard multi-image denoising technique.
  \item The interleaved flash/no-flash bursts allows us to compare to existing work in flash/no-flash fusion, both visible and dark.
  \item Because our scenes are largely static, the correspondence between the two viewpoints is the same across all acquired pairs of images. This let us use the depth map recovered from a no-flash image pair to register a subsequent flash/no-flash image pair.
\end{itemize}

If our hardware matched the idealized setup in Figure~\ref{fig:setup}(a), we could estimate a per-pixel registration across our two cameras by simply applying a stereo technique to the green channels of our two images.
However, because \camtwo's green channel is in fact quite sensitive to our NIR flash, the green channel of a flash-on image is unsuitable for stereo matching with the green channel of \camone. We circumvent this issue by computing our depth maps using the previous pair of frames, where the flash is off.
Because images in our bursts are taken in rapid succession, these depth maps tend to hold well across consecutive frames --- even when presented with subtle motions in our bursts of human subjects.
The problem of trans-modal stereo correspondence is an interesting direction for future research.

After acquiring a burst, we post-process the high-resolution raw data obtained from the two cameras to make the images amenable to joint denoising. First, we demosaic the images using method of \cite{malvar2004high} into linear raw images. These images are then downsampled to $512\times 512$. We compute a registration across the two images using the ``bilateral flow'' algorithm of \cite{Anderson2016, barron2016fast}, which produces clean, edge-aware image alignments that have been demonstrated to work well for computational photography tasks. We use optical flow in this way to circumvent the tedious calibration required by the traditional approach of rectification and stereo depth estimation.

\section{Algorithm for Low-light Imaging}

\begin{figure}[b]
  \centering
  \includegraphics[width=\linewidth]{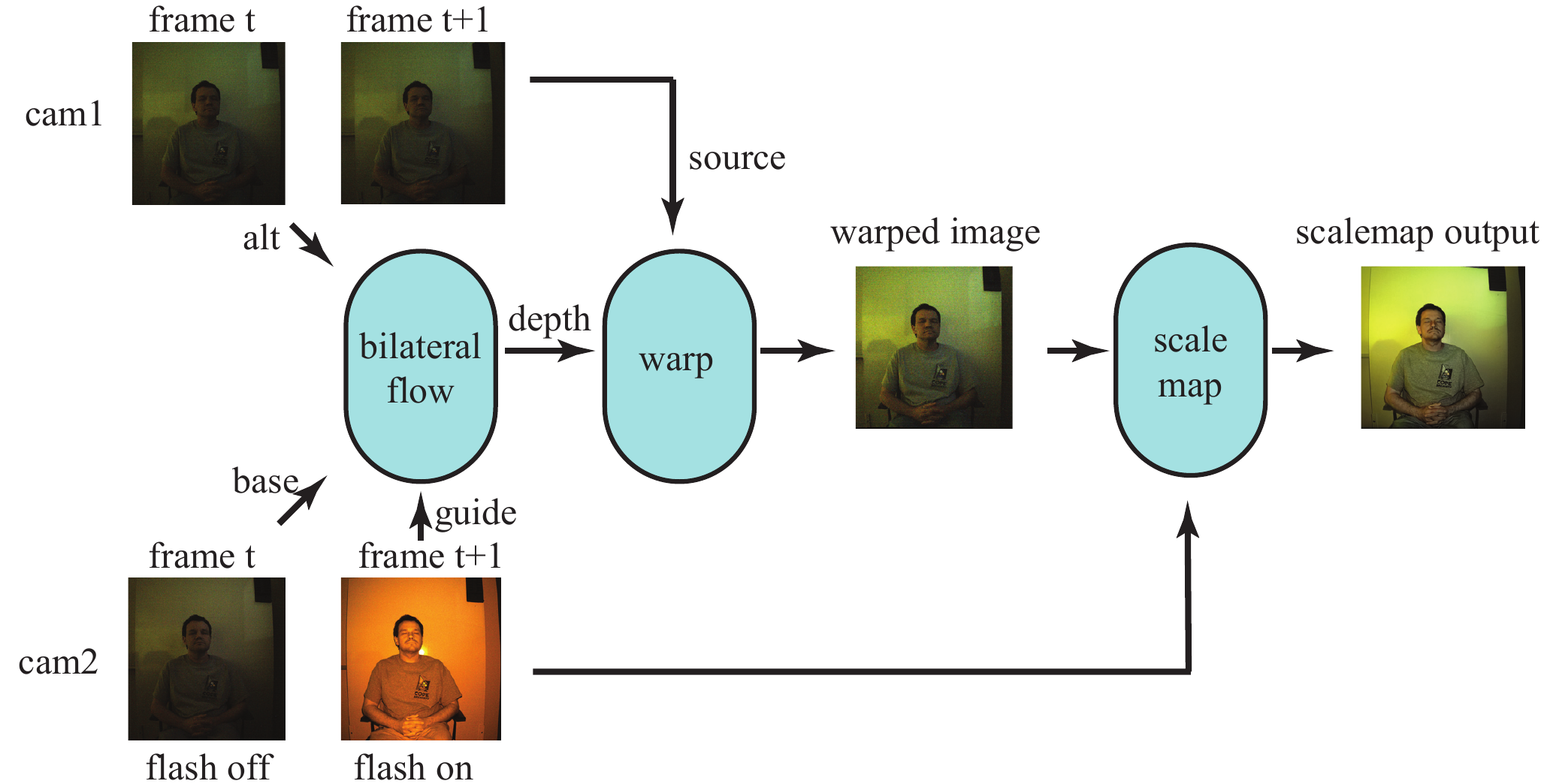}
  \caption{Our image registration pipeline.}
  \label{fig:registration}
\end{figure}

This section describes our procedure for fusing a dark flash stereo pair into a high-quality result, as shown in Figures~\ref{fig:registration} and~\ref{fig:architecture}. We first compute a per-pixel registration for the pair using \camtwo as the base, and then warp \camone accordingly. With this warped image we produce an initial fused result using the scale map algorithm of \cite{shen2015multispectral}. Because scale map fusion does not correct for all the tonal and spectral properties of our dark flash image, we then feed the initial fused image along with the warped \camone image to our neural network to produce the final image.
% Als
\subsection{Registration}
Our imaging configuration requires a per-pixel mapping between the RGB image from \camone and the flash image from \camtwo. Intuitively, we want to preserve the sharp, high-frequency details of the flash image while propagating tonal information from the RGB image, which is lower frequency and more tolerant to error. Therefore, we leave the \camtwo flash image stationary (the \base), and compute a flow field that gathers from the \camone RGB image (the \alt). We employ a variant of the ``bilateral flow'' algorithm of \cite{Anderson2016}, based on the bilateral solver of \cite{barron2016fast}, to register our stereo pair.

%In our experiments, we found existing methods~\cite{shen2014multi,kim2015dasc,zhideep} to work poorly since the images produced by the two modalities look significantly different. Instead, we employ a variant of the ``bilateral flow'' algorithm of \cite{Anderson2016}, based on the bilateral solver of \cite{barron2016fast}, to register our stereo pair.

Standard bilateral flow takes as input \base and \alt, performs tile matching to compute its data term, and optimizes for a flow field that is smooth while respecting the edges of \emph{base}. We modify standard bilateral flow so that the solution respects the edges of a third \guide image, and compute tile matching on only the green channels of \base and \alt.

If we had access to our ideal camera, the \camtwo flash image would serve as both \base and \guide. To simulate this with the burst dataset captured by our prototype camera, we first compute tile matching on the stereo pair at time $t$ when the flash is off (approximating pure green channels). We then estimate an edge-aware flow field using a bilateral solver, where as the \guide we use the image produced by \camtwo at time $t+1$ (when the flash is on). Finally, we use this flow field to warp both the RGB image at $t+1$ and the \camone long-exposure ground truth to the viewpoint of \camtwo. Figure~\ref{fig:registration} illustrates the data flow.

\subsection{Learned Image Fusion}

\label{sec:image_fusion}
After registration, for each scene and exposure setting, we have four images from the perspective of \camtwo:
\begin{enumerate}
 \item A flash image captured by \camtwo.
 \item An RGB image warped from \camone.
 \item A long-exposure RGB image warped from \camone.
 \item A long-exposure RGB image captured with \camtwo.
\end{enumerate}

% One challenge of fusing a IR-UV flash images with a non-flash image is that IR or UV flashes may change experience of high-frequency texture, or introduce new texture. Those high-frequency texture are often corrupted by noise in non-flash images (3), making it hard to determine whether to preserve them in the 
The fusion algorithm, given only 1) and 2), needs to generate an image that is close to 4).
This is nontrivial, as the flash image 1) looks significantly different from the noisy RGB image 2).
Scalemap~\cite{shen2015multispectral}, a representative state-of-the-art algorithm for cross-domain image fusion, can effectively remove much of this noise. However, it also changes the color tone and local contrast of images captured by \camone, resulting in unnatural ``hazy'' images. Skin in particular looks ``waxy'' (see Figure~\ref{fig:results}, column 4).
But since we collect ground-truth RGB images, we can use modern end-to-end discriminative learning techniques to eliminate these artifacts.

\begin{figure}[ht]
  \centering
  \includegraphics[width=\linewidth]{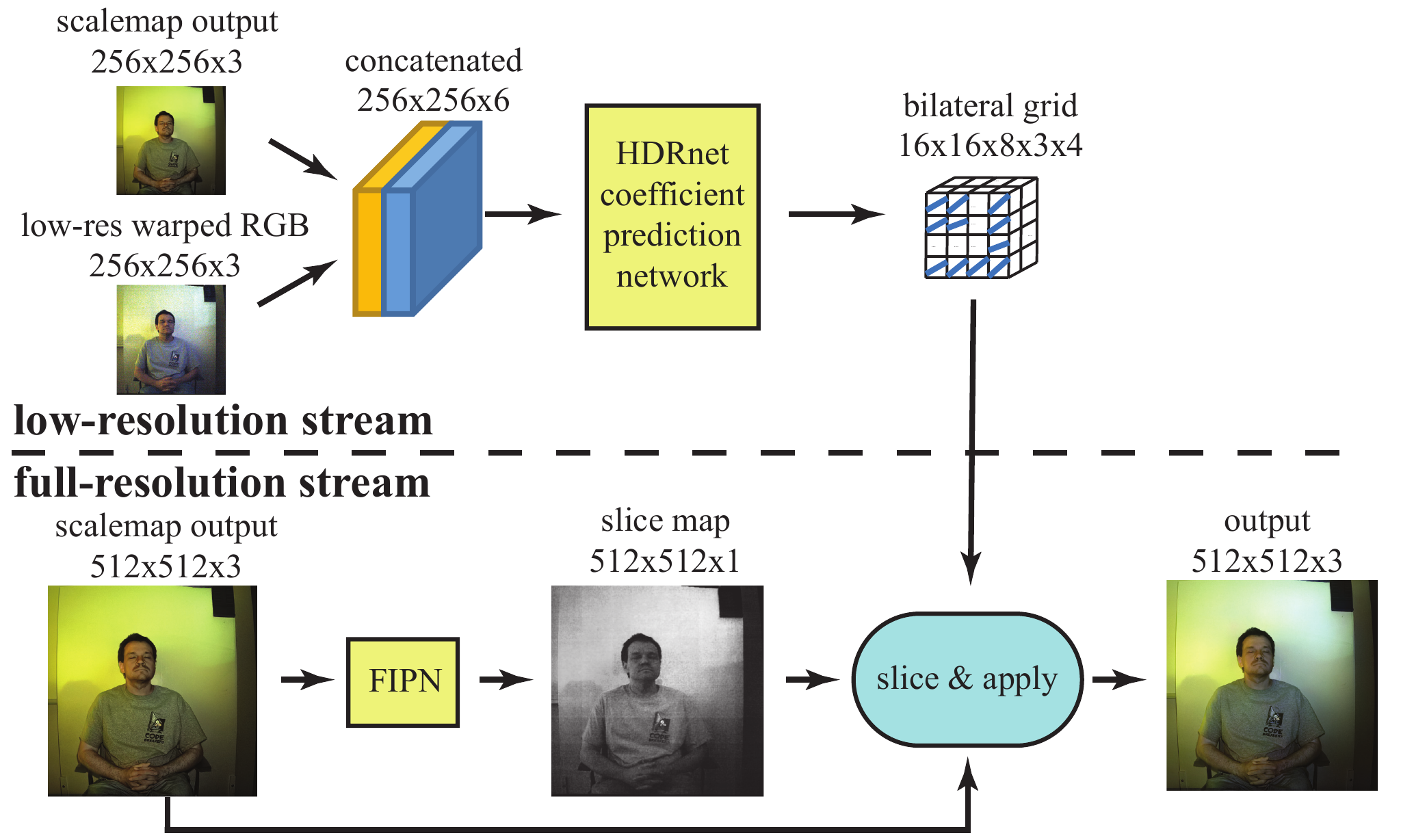}
  \vspace{-4pt}
  \caption{Our tone correction CNN architecture. Trainable blocks are rendered in yellow and optimized jointly.}
  \label{fig:architecture}
\end{figure}

One may be tempted to train a general end-to-end model that, from inputs 1) and 2), synthesizes an output resembling 4). However, this places a large burden on the network as it must learn to simultaneously denoise, account for mis-registration, and correct for flash shadows in addition to color tone correction, all with a limited dataset. Indeed, our experiments with this approach failed to produce promising results (as shown by the ``FIPN'' entry in Figure~\ref{fig:results}). We instead make learning easier by asking the network to only learn color and contrast correction. 

Our network takes as input the Scalemap algorithm's output, as these images already have low noise and simply require tonal correction. To ensure that the network exclusively corrects color, and that training does not fail due to stereo misalignment across the input images, for training we use as ``ground truth'' image 3): the  long-exposure RGB image warped from \camone. This does mean, however, that geometric errors from stereo registration will persist in the output of our model.

% Text written by Kevin:
%However, it occasionally results in some artifacts like ``hazy'' images or a ``waxy'' appearance on human skin, and often has trouble coping with the shadow cast by our flash, which it was not designed to correct\jc{see Figure, column ?}.
%Modifying an optimization framework like scalemap to address these issues is challenging, due to its non-learned hand-engineered nature.
%But because our dataset acquisition process includes the acquisition of a ground-truth RGB image, we can use modern end-to-end discriminative learning techniques to eliminate these artifacts.
% Despite these moderate shortcomings, scalemap significantly outperformed other methods (flash/noflash, guided filter, muGIF),
% Despite these moderate shortcomings, scalemap produces a low-noise image that is a good starting point for training a neural network that can remove the remaining artifacts.

Our network is based on ``HDRnet''~\cite{GharbiSIGGRAPH2017}, a deep neural network that predicts edge-aware local and global tone correction functions (Figure~\ref{fig:architecture}). It consists of a low-resolution stream that predicts an image transformation encoded as an \emph{affine bilateral grid} (a bilateral grid where each cell contains an affine transformation of RGB values), and a full-resolution stream that learns how to best \emph{slice} into the grid and apply the resulting transformation, which is then used to produce the output image.

A straightforward adaptation of HDRnet to our dataset would be to use as input both the flash image 1), the output of Scalemap applied to 1), and the no-flash RGB image 2), all concatenated together as a single 6-channel image, and to modify the bilateral grid to contain $3 \times 7$ affine transformations accordingly (Figure~\ref{fig:architecture}, top). However, we found this to work poorly because HDRnet is unable to express a bilateral grid of affine transformations that removes the shadows cast by the dark flash, in part because HDRnet is designed to ``slice'' from this grid using just per-pixel luma.

To address this issue, we generalize the model by using another deep network that learns \emph{how to slice} into the bilateral grid (Figure~\ref{fig:architecture}, bottom). We replace HDRnet's simple per-pixel network with a significantly more expressive 9-layer fully-convolutional network modeled after the Fast Image Processing Network (FIPN) of \cite{Qifeng2017}, to produce a \emph{slice map}. Though the FIPN model is quite general, because we use the output of this model only to slice from a bilateral grid (instead of using it to synthesize a complete image) our HDRnet-like architecture still constrains the output of our complete model to be a local affine transformation of the input image.
To accommodate our noisy inputs, we downsample our noisy RGB and Scalemap outputs and concatenate them as input to HDRnet's low-resolution stream. At full-resolution, we predict the slice map using only the Scalemap output. We then slice and apply local $3 \times 4$ affine transformations to each pixel. We found that adding the noisy RGB image to the full-resolution stream did not appreciably improve performance. Finally, we replace the luma channel of HDRnet's output with the luma channel from Scalemap, which helps to preserve some details.

After visual inspection, we decided to conduct all experiments only on the $T/5$ subset as it most closely mimics the noise characteristics of modern smartphone cameras. Furthermore, after preliminary experiments, we decided to %forego using NIR+NUV images due to the properties of UV light \barron{This makes it sound like we're not using NIR or NUV. Clarify that we drop NUV and use only NIR}. 
not use NUV flash and use only NIR. Many materials such as fabric, paper, and glass \emph{fluoresce}--that is, when they absorb UV light, they re-emit that light's energy as visible blue light~\cite{lakowiczprinciples}. Although NUV can clearly help with denoising and many materials such as human skin do not fluoresce, the inconsistencies that fluorescence caused in our dataset made it challenging to achieve good results. The supplement contains one scene depicting fluorescence.

We randomly select 90\% of our dataset for training, leaving the rest for testing. Since both HDRnet and FIPN are resolution-independent, we downsample our high-resolution images to $256 \times 256$ to accelerate training, but evaluate our results at $512 \times 512$. As a baseline, we also train FIPN at $512 \times 512$. For all networks, we trained on batches of size 4 using the Adam optimizer~\cite{AdamOpt} with learning rate $10^{-4}$, $\beta_1=0.9$, and $\beta_2=0.999$, for $500$ epochs.
Our model is implemented in TensorFlow.

% \barron{We have to explain what our error measures are. What is ``Gram''? Cite it?} Tianfan: addressed this one the result section.

% network is a balance. we use as hdrnet \jc{cite gharbi} as the basis.
% don't significantly alter input.

% \tianfan{This paragraph is not finished yet} To ease learning, we first use the scale map algorithm to combine the noisy image (captured by \camone) and the flash image (captured by \camtwo) registered to \camone. Then, we then apply HDRnet \tianfan{citation} to bring the local contrast and colortone of the scalemap output closer to the long-expose image without flash. 

% training deep neural networks that can directly synthesize 

% To make machine learning easier, we first apply the scale map algorithm \cite{shen2015multispectral} to the pair to produce a decent RGB image. This removes the noise and reduces it to a problem of removing flash shadows and tone mapping - problems that have yet to be tackled by traditional image processing and a good candidate for modern machine learning.
% Since we have ground truth, the long exposure image, we can learn a mapping from the pair (warped RGB image, scale map based method output) to ground truth. The entire processing pipeline is the following:

%\subsection{Tone adjustment}

% \subsection{Bokeh effect}

\section{Results}

\begin{figure*}
	\center
	\includegraphics[width=1\textwidth]{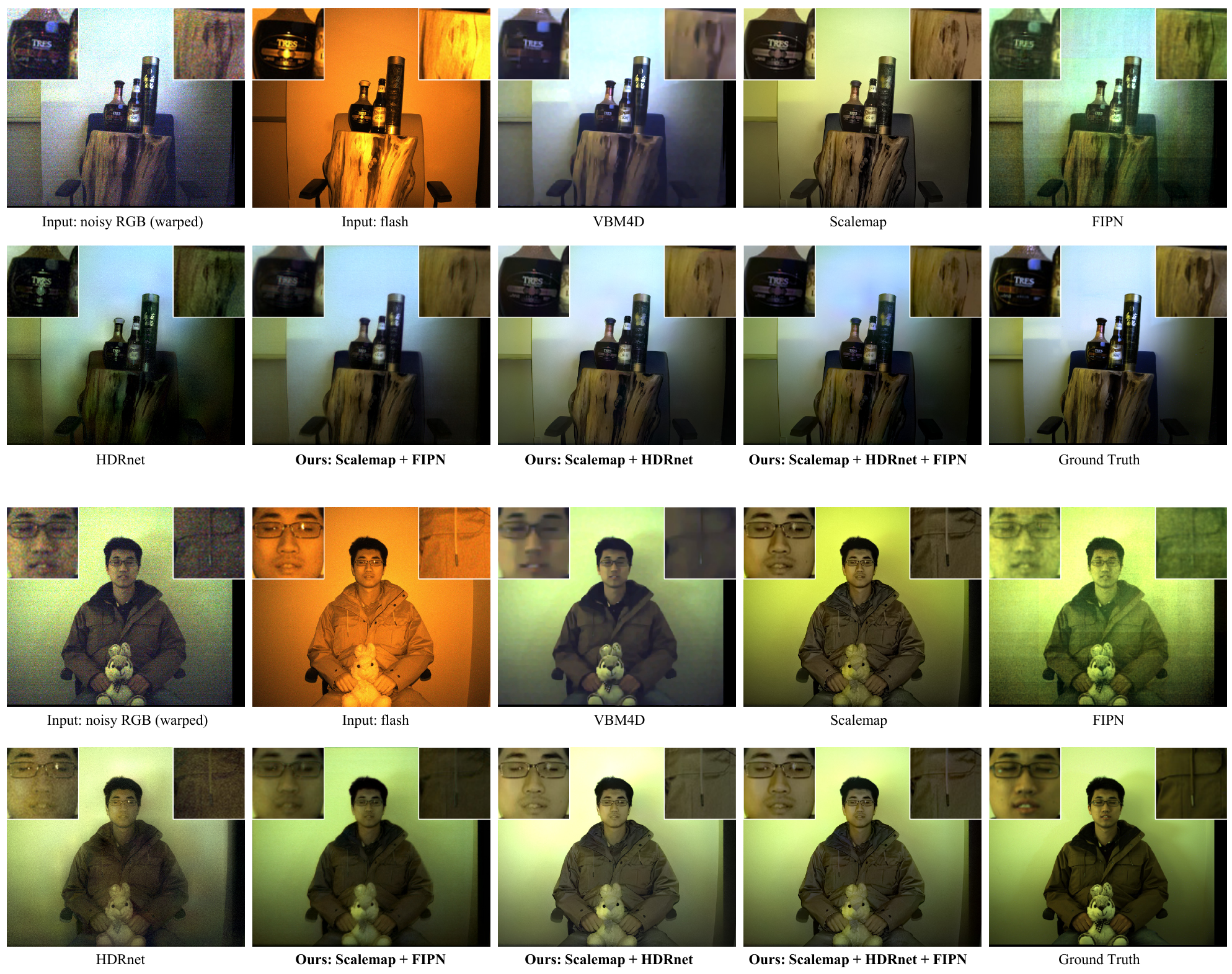}\\
	\caption{Our results compared with previous methods and our ground truth on two scenes from our test set. The inputs to our system are a noisy RGB image from \camone (visualized here with a digital gain of $\times 5$ for the sake of visualization) and a dark flash image from \camtwo. VBM4D denoises a burst of four noisy RGB images. Scalemap~\cite{shen2015multispectral} fuses the RGB and dark flash images, which removes much of the noise but also results in a color shift and poor local contrast. Direct image synthesis using FIPN reveals significant artifacts from dilated convolution, while unmodified HDRnet can recover the global color tone but fails to improve local contrast over Scalemap. Our neural network restores this detail, resulting in an image substantially close to the ground truth.}
	\label{fig:results}
\end{figure*}

\begin{figure*}
	\center
	\includegraphics[width=1\textwidth]{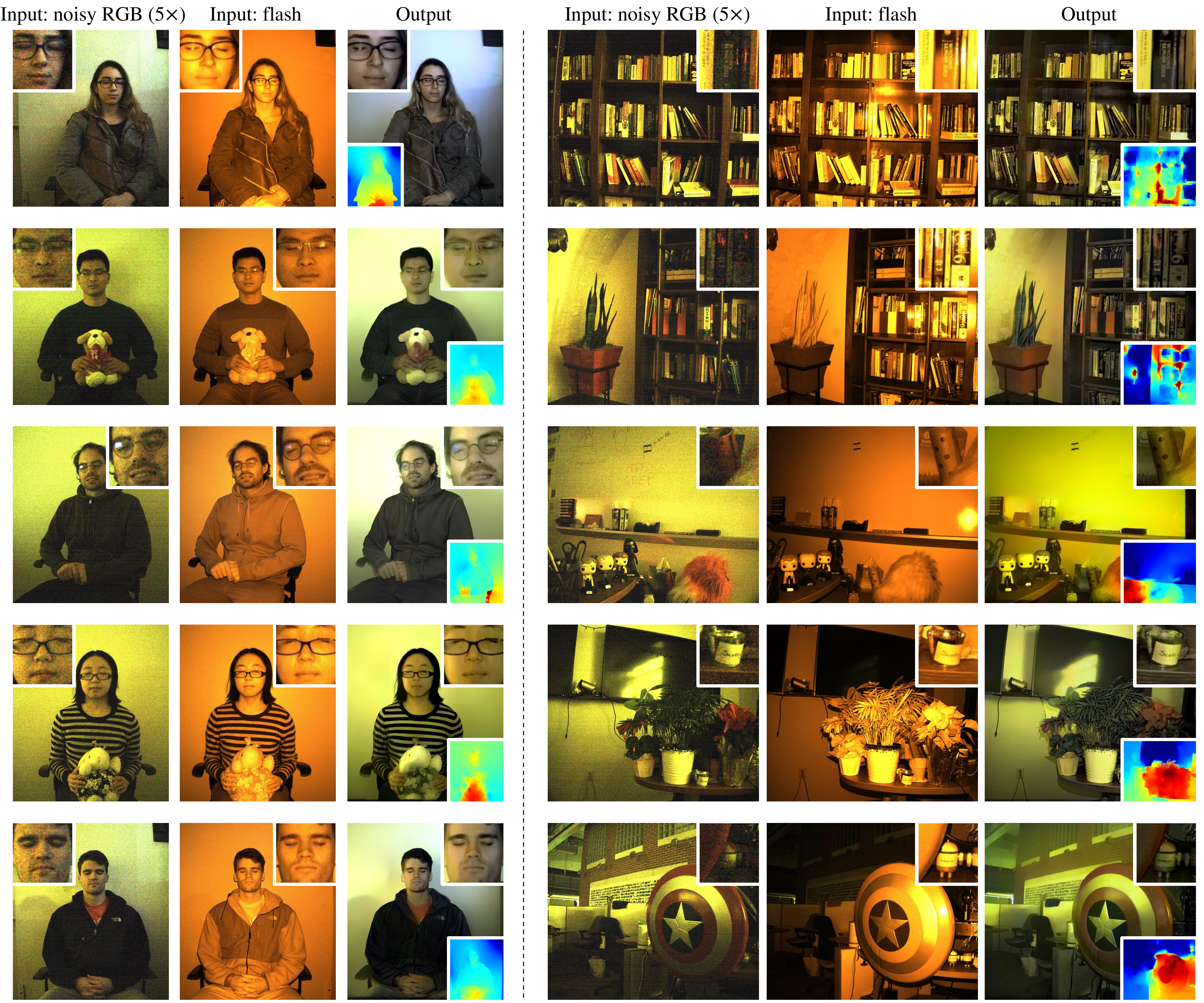}\\
	\caption{Inputs and our results on several scenes of both human and still-life subjects. Images are best viewed zoomed-in on a computer where noise is visible.}
	\label{fig:results_all}
\end{figure*}

\begin{table}[b]
    \begin{center}
    \begin{tabular}{l | ccc }
 & \multicolumn{3}{c}{Error Metrics} \\
Method & PSNR & SSIM & Style \\ \hline
% VGG loss
% Noisy input captured by \camone & $20.37$ & $0.35$ & $77.13$\\
% BM3D & $22.08$ & $0.79$ & $33.96$\\
% VBM4D & $22.23$ & $0.79$ & $\mathbf{33.35}$\\
% Scalemap \cite{shen2015multispectral} & $20.31$ & $0.68$ & $43.93$\\
% HDRnet \cite{GharbiSIGGRAPH2017} & $22.27$ & $0.58$ & $56.63$\\
% FIPN \cite{Qifeng2017} & $21.54$ & $0.71$ & $50.33$\\
% \hline
% Ours: Scalemap + FIPN & $\mathbf{30.12}$ & $\mathbf{0.83}$ & $39.87$\\
% Ours: Scalemap + HDRnet & $27.46$ & $0.81$ & $33.60$\\
% Ours: Scalemap + HDRnet + FIPN & $27.53$ & $0.82$ & $33.39$\\
% Gram loss
Noisy input captured by \camone & $20.37$ & $0.35$ & $4.07$\\
BM3D & $22.08$ & $0.79$ & $1.97$\\
VBM4D & $22.23$ & $0.79$ & $1.81$\\
Scalemap \cite{shen2015multispectral} & $20.31$ & $0.68$ & $1.74$\\
% Scalemap on White Flash & $16.43$ & $0.66$ & $2.15$\\
HDRnet \cite{GharbiSIGGRAPH2017} & $22.27$ & $0.58$ & $2.89$\\
\hline
FIPN \cite{Qifeng2017} & $21.54$ & $0.71$ & $2.32$\\
Ours: Scalemap + FIPN & $\mathbf{27.66}$ & $\mathbf{0.80}$ & $2.08$\\
Ours: Scalemap + HDRnet & $23.16$ & $0.74$ & $1.62$\\
Ours: Scalemap + HDRnet + FIPN & $24.47$ & $0.75$ & $\mathbf{1.43}$\\
    \end{tabular}%
    \vspace{3pt}
    \caption{Quantitative results under three error metrics, comparing variants of our method against a number of baseline techniques. For PSNR and SSIM, higher is better. For ``Style'', lower is better.
    }
    \label{tab:results}
    \end{center}
    \vspace{-15pt}
\end{table}

% \begin{table}[b]
%     \begin{center}
%     \begin{tabular}{l | ccc }
%  & \multicolumn{3}{c}{Error Metrics} \\
% Method & PSNR & SSIM & Style \\ \hline
% Noisy input captured by \camone & $17.02$ & $0.2449$ & $3.952$ \\
% BM3D \cite{BM3D} & $18.65$ & $0.6844$ & $7.304$ \\
% VBM4D \cite{VBM4D} & $18.97$ & $0.6807$ & $7.298$ \\
% Scalemap \cite{shen2015multispectral} & 18.34 & $0.6715$ & $6.958$ \\
% HDRnet \cite{GharbiSIGGRAPH2017} & $20.36$ & $0.4982$ & $2.840$ \\
% FIPN \cite{Qifeng2017} & $22.84$ & $0.7902$ & $3.663$ \\
% \hline
% Ours: Scalemap + FIPN & $\mathbf{23.47}$ & $\mathbf{0.7822}$ & $2.948$ \\
% Ours: Scalemap + HDRnet & 17.96 & $0.6757$ & $5.237$ \\
% Ours: Scalemap + HDRnet + FIPN & $21.34$ & $0.7013$ & $\mathbf{2.828}$ \\ 
%     \end{tabular}%
%     \vspace{3pt}
%     \caption{Quantitative results under three error metrics, comparing variants of our method against a number of baseline techniques. For PSNR and SSIM, higher is better. For ``Style'', lower is better.
%     }
%     \label{tab:results}
%     \end{center}
%     \vspace{-15pt}
% \end{table}

We evaluate several variants of our method by comparing it to a number of existing baselines:
\begin{itemize}
\item {\bf Input}: we report the error of the noisy RGB image from \camone compared to the long exposure image, as a point of reference.
\item {\bf BM3D}~\cite{BM3D}: a classical single-image denoising algorithm. We use the last no-flash frame from \camone (the RGB camera) as input.
\item {\bf VBM4D}~\cite{VBM4D}: a multi-image denoising algorithm. We use all $4$ no-flash images from \camone (the RGB camera) as input.
\item {\bf HDRnet}~\cite{GharbiSIGGRAPH2017}: a fast and effective neural network model for learning local tone-mapping operators. We directly apply HDRnet to a concatenation of the registered RGB image captured by \camone and the flash image captured by \camtwo, training the network to approximate the long-exposure RGB image warped from \camone.
\item {\bf FIPN}~\cite{Qifeng2017}: a general purpose neural network model for arbitrary imaging transformations. For training, we use the same input/output images as in our HDRnet baseline.
\item {\bf Scalemap}~\cite{shen2015multispectral}: an optimization algorithm designed for fusing RGB and hyperspectral images. We use the registered RGB image captured by \camone and the flash image captured by \camtwo as input.
\end{itemize}
% \item {\bf Scalemap on White Flash}: similar to Scalemap, but instead of noisy RGB image with a white-flash image, instead of IR flash (dark flash).  #tianfan: remove white flash
% \end{itemize}

The algorithm we described in Section~\ref{sec:image_fusion} (which we alternatingly refer to ``Scalemap + HDRnet + FIPN'', or just ``our model'') is constructed out of several of our baseline algorithms and trained end-to-end, which we found to produce the most visually pleasing results. We additionally evaluate against two ablations of our model, % that use prior work as building blocks in different ways
both of which takes the concatenation of the output by Scalemap and noisy RGB image as input:
% \barron{These baselines don't clarify that Scalemap's output is used as input to each model. I also found both descriptions much more confusing than the latex comments after them.} % tianfan: Add one sentence to the end of this paragraph to clarify that.
\begin{itemize}
\item {\bf Scalemap + HDRnet}: Instead of using the output of an FIPN model as the guide map in HDRnet, we use the simple trainable piecewise linear functions originally proposed by Gharbi et al~\shortcite{GharbiSIGGRAPH2017}.% This model is effectively just HDRnet applied to the output of Scalemap.
\item {\bf Scalemap + FIPN}: Instead of using HDRnet to produce an output image, which restricts the model to only being able to estimate local affine transformations, we train FIPN to directly estimate the output image.% using the concatenation of the output of Scalemap~\cite{shen2015multispectral} and the noisy RGB image captured by \camone.
\end{itemize}
% Besides, we also compare the output merging noisy RGB image with a white-flash image (instead of UV or IR flash). We still Scalemap to fuse these two images and refers to this method as {\bf Scalemap on White Flash}.

To quantitatively compare the results of our various baselines and model variants to our long-exposure ground truth images, we first rescale outputs to have the same brightness (average RGB value) of the long-exposure ground truth to account for any brightness variation. Then we use three evaluation metrics: PSNR, SSIM~\cite{Wang04SSIM}, and a perceptual error metric (called  ``Style'' in our table). PSNR simply measures any per-pixel differences, while SSIM focuses more on structural differences and is invariant to errors that can be modeled as local shifts or scales. Both of these measures are sensitive to small misalignments in their input images, and because such misalignments are common in our dataset, neither metric is well suited to our task. For this reason, we use the perceptual metric of~\cite{gatys2016image,Johnson2016Perceptual}, which is based on \emph{texture similarity} and more forgiving to misalignments. It is commonly used for style transfer applications and is based on the Gram matrix of the feature activations (we used conv2, conv3, and conv4) of a pretrained VGG-16~\cite{simonyan2014very} image classification network.

Table~\ref{tab:results} contains a quantitative evaluation on our test set, and Figure~\ref{fig:results} contains a zoomed-in comparison of two examples.
The single- and multi-frame techniques for denoising the RGB image(s) without the aid of of the dark flash image tend to generate blurry or oversmoothed output, and as such have lower PSNRs than all other techniques, which do use the dark flash image.
Scalemap generates sharp images, but introduces obvious color shifts compared to the ground truth.
All three non-learning approaches (BM3D, VBM4D, and Scalemap) have the lowest PSNR in all the approaches, demonstrating the value of learning for this task.
Our two learning-based baselines, HDRnet and FIPN, achieve higher PSNR than the non-learning-based approaches, but both of them introduce a significant amount of noise to the output image (Figure~\ref{fig:results}).

Our ``Scalemap + FIPN'' ablation achieves the highest PSNR and SSIM. However, upon inspection, we see that its output images are slightly blurred, perhaps due to insufficient, or perhaps due to the fact that the transformation cannot be entirely modeled by the strong locally-affine constraint enforced by HDRnet (see Figure~\ref{fig:results}). The ``Style'' metric appears to be sensitive to the artifacts produced by this model and penalizes it accordingly.
Both the ``Scalemap + HDRnet'' ablation and our complete ``Scalemap + HDRnet + FIPN'' model successfully preserve the color, tone, and contrast of RGB images, while removing most of the high-frequency noise. Our ``Scalemap + HDRnet + FIPN'' model better preserves local contrast than its ``Scalemap + HDRnet'' ablation, as can be seen in the texture of the wooden structure in the bottom example of Figure~\ref{fig:results}. Figure \ref{fig:results_all} contains additional results on a variety of human and still-life subjects in low-light indoor environments. Our technique produces both a low-noise RGB image as well as a dense edge-aware depth map.

\section{Conclusions}

% future work:
% real hardware
% addressing stereo issues
% figuring out when to return just the left
% merging bursts from the two cameras

% Contributions: (1) propose the new setup, where a second IR-G-UV camera is added; the  original RGB camera is untouched which is used for regular photography;
% AE algorithm, basic (+ how to find the gains in between for the bursts)
% (2) propose the low-light imaging flow: the two cameras capture images with dark flashes on, find the flow through G-channel, guided denoising, tone adjustment (a cnn to make it look nice), depth and bokeh effect (optional);
% (3) a burst dataset, in this  paper it is used to showcase the advantages of our method over previous methods like 

We have presented a design for a stereoscopic dark flash camera, which acquires stereo pairs in which one camera images the complete visible spectrum while the other camera selectively captures some visible and some hyperspectral light. When paired with a hyperspectral flash this camera configuration allows for the acquisition of ``dark flash'' images even in the presence of motion, thereby allowing for low-noise photography in low-light environments, without disturbing human subjects with a dazzling flash. To this end we have constructed a hardware prototype that approximates our idealized camera configuration and a dataset acquisition procedure that circumvents the shortcomings of our hardware prototype while also capturing ground truth long-exposure images.
With the goal of fusing our dark flash stereo pairs into a low-noise and visually pleasing image, we have presented set of novel deep neural network architectures which we train end-to-end to regress from dark flash stereo pairs to the true RGB long exposure images in our dataset. We show that these learned fused images have the low-noise properties of our dark flash image, while retaining the aesthetically pleasing tonal properties of our noisy no-flash RGB images.

% only include the acknowledgements in the camera-ready
\ifpeerreview
\else
% use section* for acknowledgment
\section*{Acknowledgment}

The authors would like to thank Nori Kanazawa, Roman Lewkow, Dilip Krishnan and Marc Levoy for the help and constructive discussions. We would like to thank our volunteers for participating in the experiments。
\fi

% references section
\bibliographystyle{IEEEtran}
\bibliography{bibliography}

% Generated by IEEEtran.bst, version: 1.14 (2015/08/26)
\begin{thebibliography}{10}
\providecommand{\url}[1]{#1}
\csname url@samestyle\endcsname
\providecommand{\newblock}{\relax}
\providecommand{\bibinfo}[2]{#2}
\providecommand{\BIBentrySTDinterwordspacing}{\spaceskip=0pt\relax}
\providecommand{\BIBentryALTinterwordstretchfactor}{4}
\providecommand{\BIBentryALTinterwordspacing}{\spaceskip=\fontdimen2\font plus
\BIBentryALTinterwordstretchfactor\fontdimen3\font minus
  \fontdimen4\font\relax}
\providecommand{\BIBforeignlanguage}[2]{{%
\expandafter\ifx\csname l@#1\endcsname\relax
\typeout{** WARNING: IEEEtran.bst: No hyphenation pattern has been}%
\typeout{** loaded for the language `#1'. Using the pattern for}%
\typeout{** the default language instead.}%
\else
\language=\csname l@#1\endcsname
\fi
#2}}
\providecommand{\BIBdecl}{\relax}
\BIBdecl

\bibitem{hasinoff2016burst}
S.~W. Hasinoff, D.~Sharlet, R.~Geiss, A.~Adams, J.~T. Barron, F.~Kainz,
  J.~Chen, and M.~Levoy, ``Burst photography for high dynamic range and
  low-light imaging on mobile cameras,'' \emph{SIGGRAPH}, 2016.

\bibitem{petschnigg2004digital}
G.~Petschnigg, R.~Szeliski, M.~Agrawala, M.~Cohen, H.~Hoppe, and K.~Toyama,
  ``Digital photography with flash and no-flash image pairs,'' \emph{ACM TOG},
  2004.

\bibitem{eisemann2004flash}
E.~Eisemann and F.~Durand, ``Flash photography enhancement via intrinsic
  relighting,'' \emph{ACM TOG}, 2004.

\bibitem{krishnan2009dark}
D.~Krishnan and R.~Fergus, ``Dark flash photography,'' \emph{SIGGRAPH}, 2009.

\bibitem{barron2015fast}
J.~T. Barron, A.~Adams, Y.~Shih, and C.~Hern{\'a}ndez, ``Fast bilateral-space
  stereo for synthetic defocus,'' \emph{CVPR}, 2015.

\bibitem{bayer1976color}
B.~E. Bayer, ``Color imaging array,'' 1976, {US} Patent 3,971,065.

\bibitem{malvar2004high}
H.~S. Malvar, L.-w. He, and R.~Cutler, ``High-quality linear interpolation for
  demosaicing of bayer-patterned color images,'' \emph{ICASSP}, 2004.

\bibitem{jiang2013space}
J.~Jiang, D.~Liu, J.~Gu, and S.~S{\"u}sstrunk, ``What is the space of spectral
  sensitivity functions for digital color cameras?'' \emph{WACV}, 2013.

\bibitem{litwiller2001ccd}
D.~Litwiller, ``Ccd vs. cmos,'' \emph{Photonics spectra}, 2001.

\bibitem{dabov2007image}
K.~Dabov, A.~Foi, V.~Katkovnik, and K.~Egiazarian, ``Image denoising by sparse
  3-d transform-domain collaborative filtering,'' \emph{IEEE TIP}, 2007.

\bibitem{elad2006image}
M.~Elad and M.~Aharon, ``Image denoising via sparse and redundant
  representations over learned dictionaries,'' \emph{IEEE TIP}, 2006.

\bibitem{gu2014weighted}
S.~Gu, L.~Zhang, W.~Zuo, and X.~Feng, ``Weighted nuclear norm minimization with
  application to image denoising,'' \emph{CVPR}, 2014.

\bibitem{chen2018learSeeDark}
C.~Chen, Q.~Chen, J.~Xu, and V.~Koltun, ``Learning to see in the dark,''
  \emph{CVPR}, 2018.

\bibitem{fast-burst-images-denoising}
Z.~Liu, L.~Yuan, X.~Tang, M.~Uyttendaele, and J.~Sun, ``Fast burst images
  denoising,'' \emph{SIGGRAPH Asia}, 2014.

\bibitem{Dabov2007}
K.~Dabov, A.~Foi, and K.~Egiazarian, ``Video denoising by sparse 3d
  transform-domain collaborative filtering,'' \emph{Eusipco}, 2007.

\bibitem{heide2014flexisp}
F.~Heide, M.~Steinberger, Y.-T. Tsai, M.~Rouf, D.~Pająk, D.~Reddy, O.~Gallo,
  J.~L. abd Wolfgang~Heidrich, K.~Egiazarian, J.~Kautz, and K.~Pulli,
  ``Flexisp: A flexible camera image processing framework,'' \emph{SIGGRAPH
  Asia}, 2014.

\bibitem{Fergus06}
R.~Fergus, B.~Singh, A.~Hertzmann, S.~T. Roweis, and W.~T. Freeman, ``Removing
  camera shake from a single photograph,'' \emph{SIGGRAPH}, 2006.

\bibitem{he2010guided}
K.~He, J.~Sun, and X.~Tang, ``Guided image filtering,'' 2010.

\bibitem{shen2015multispectral}
X.~Shen, Q.~Yan, L.~Xu, L.~Ma, and J.~Jia, ``Multispectral joint image
  restoration via optimizing a scale map,'' \emph{IEEE TPAMI}, 2015.

\bibitem{guo2017mutually}
X.~Guo, Y.~Li, and J.~Ma, ``Mutually guided image filtering,'' \emph{ACM
  Multimedia}, 2017.

\bibitem{spooren2016rgb}
N.~Spooren, B.~Geelen, K.~Tack, A.~Lambrechts, M.~Jayapala, R.~Ginat, Y.~David,
  E.~Levi, and Y.~Grauer, ``Rgb-nir active gated imaging,''
  \emph{Electro-Optical and Infrared Systems: Technology and Applications
  XIII}, 2016.

\bibitem{tack2012compact}
N.~Tack, A.~Lambrechts, P.~Soussan, and L.~Haspeslagh, ``A compact, high-speed,
  and low-cost hyperspectral imager,'' \emph{Silicon Photonics VII}, 2012.

\bibitem{geelen2015tiny}
B.~Geelen, C.~Blanch, P.~Gonzalez, N.~Tack, and A.~Lambrechts, ``A tiny vis-nir
  snapshot multispectral camera,'' \emph{Advanced Fabrication Technologies for
  Micro/Nano Optics and Photonics VIII}, 2015.

\bibitem{Anderson2016}
R.~Anderson, D.~Gallup, J.~T. Barron, J.~Kontkanen, N.~Snavely, C.~Hern\'andez,
  S.~Agarwal, and S.~M. Seitz, ``Jump: Virtual reality video,'' \emph{SIGGRAPH
  Asia}, 2016.

\bibitem{barron2016fast}
J.~T. Barron and B.~Poole, ``The fast bilateral solver,'' \emph{ECCV}, 2016.

\bibitem{GharbiSIGGRAPH2017}
M.~Gharbi, J.~Chen, J.~T. Barron, S.~W. Hasinoff, and F.~Durand, ``Deep
  bilateral learning for real-time image enhancement,'' \emph{SIGGRAPH}, 2017.

\bibitem{Qifeng2017}
Q.~Chen, J.~Xu, and V.~Koltun, ``Fast image processing with fully-convolutional
  networks,'' \emph{ICCV}, 2017.

\bibitem{lakowiczprinciples}
J.~Lakowicz, ``Principles of fluorescence spectroscopy,'' 1999.

\bibitem{AdamOpt}
D.~P. Kingma and J.~Ba, ``Adam: A method for stochastic optimization,''
  \emph{CoRR}, vol. abs/1412.6980, 2014.

\bibitem{BM3D}
K.~Dabov, A.~Foi, V.~Katkovnik, and K.~Egiazarian, ``Bm3d image denoising with
  shape-adaptive principal component analysis,'' \emph{SPARS}, 2009.

\bibitem{VBM4D}
M.~Maggioni, G.~Boracchi, A.~Foi, and K.~Egiazarian, ``Video denoising,
  deblocking, and enhancement through separable 4-d nonlocal spatiotemporal
  transforms,'' \emph{IEEE TIP}, 2012.

\bibitem{Wang04SSIM}
Z.~Wang, A.~C. Bovik, H.~R. Sheikh, and E.~P. Simoncelli, ``Image quality
  assessment: From error visibility to structural similarity,'' \emph{IEEE
  TIP}, 2004.

\bibitem{gatys2016image}
L.~A. Gatys, A.~S. Ecker, and M.~Bethge, ``Image style transfer using
  convolutional neural networks,'' 2016.

\bibitem{Johnson2016Perceptual}
J.~Johnson, A.~Alahi, and L.~Fei-Fei, ``Perceptual losses for real-time style
  transfer and super-resolution,'' \emph{ECCV}, 2016.

\bibitem{simonyan2014very}
K.~Simonyan and A.~Zisserman, ``Very deep convolutional networks for
  large-scale image recognition,'' \emph{arXiv:1409.1556}, 2014.

\end{thebibliography}

% biography section -- this is only for the camera-ready version. 
\ifpeerreview
\else
% 
% If you have an EPS/PDF photo (graphicx package needed) extra braces are
% needed around the contents of the optional argument to biography to prevent
% the LaTeX parser from getting confused when it sees the complicated
% \includegraphics command within an optional argument. (You could create
% your own custom macro containing the \includegraphics command to make things
% simpler here.)
%\begin{IEEEbiography}[{\includegraphics[width=1in,height=1.25in,clip,keepaspectratio]{mshell}}]{Michael Shell}
% or if you just want to reserve a space for a photo:

\begin{IEEEbiography}[{\includegraphics[width=1in,height=1.25in,clip,keepaspectratio]{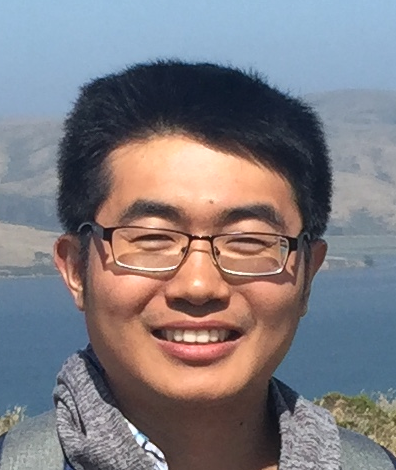}}]{Jian Wang}
	is a research scientist at Snap Inc., working on computational photography and computer vision. He received his Ph.D. in ECE from CMU, working with Aswin Sankaranarayanan and Srinivasa Narasimhan. Before that, he received his M.S. degree from University of Science and Technology of China, and his B.Eng degree and B.Ec. degree from Xi'an Jiao Tong University. He is specifically interested in 3D reconstruction and high-performance imaging.
\end{IEEEbiography}

\begin{IEEEbiography}[{\includegraphics[width=1in,height=1.25in,clip,keepaspectratio]{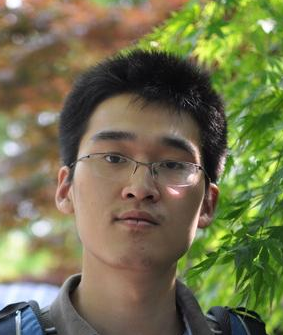}}]{Tianfan Xue}
	is a researcher at Google, working on computational photography, computer vision, and machine learning. He received his Ph.D. in EECS from MIT, working with William T. Freeman. Before that, he received his B.E. degree from Tsinghua Universtiy, and his M.Phil. degree from The Chinese University of Hong Kong. He is specifically interested in motion estimation, 3D geometry reconstruction, and machine-learning-based image and video processing.
\end{IEEEbiography}

\begin{IEEEbiography}[{\includegraphics[width=1in,height=1.25in,clip,keepaspectratio]{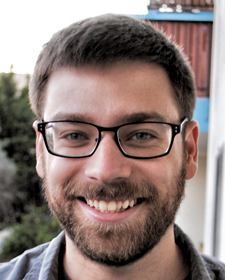}}]{Jonathan T. Barron}
	is a staff research scientist at Google, working on computer vision and computational photography. He received a PhD in Computer Science from the University of California, Berkeley in 2013, where he was advised by Jitendra Malik, and he received a Honours BSc in Computer Science from the University of Toronto in 2007. His research interests include computer vision, machine learning, computational photography, shape reconstruction, and biological image analysis. He received a National Science Foundation Graduate Research Fellowship in 2009, the C.V. Ramamoorthy Distinguished Research Award in 2013, and the ECCV Best Paper Honorable Mention in 2016.
\end{IEEEbiography}

\begin{IEEEbiography}[{\includegraphics[width=1in,height=1.25in,clip,keepaspectratio]{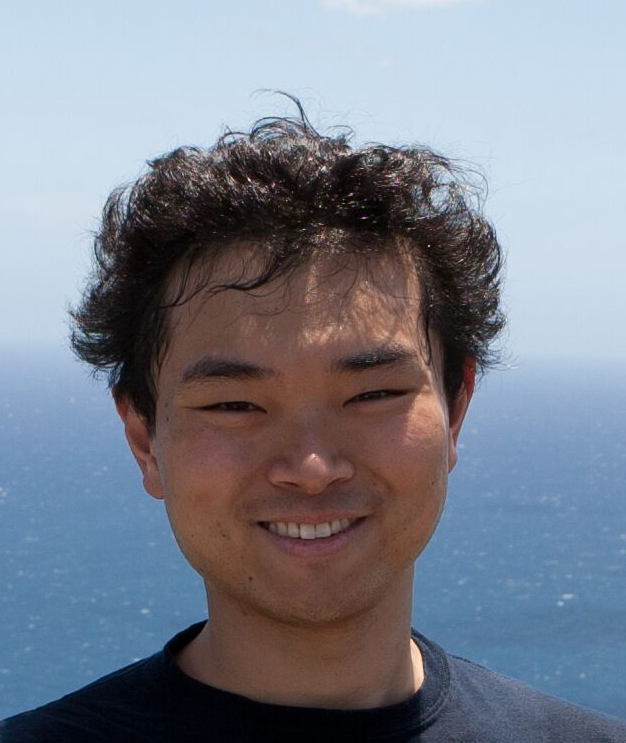}}]{Jiawen Chen}
	is a software engineer at Google Research, working on computational photography and machine learning. He received his Ph.D. in EECS from MIT CSAIL, where he was advised by Frédo Durand. Previously, he received an M.Eng. in EECS, and two S.B. degrees in EECS and Physics, also from MIT. His research interests include computational photography, computer graphics, computer vision, and interactive techniques.
\end{IEEEbiography}

% if you will not have a photo at all:
%\begin{IEEEbiographynophoto}{John Doe}
%Biography text here.
%\end{IEEEbiographynophoto}

% insert where needed to balance the two columns on the last page with
% biographies
%\newpage

%\begin{IEEEbiographynophoto}{Jane Doe}
%Biography text here.
%\end{IEEEbiographynophoto}

\fi

% that's all folks
\end{document}